\documentclass[letterpaper, 10 pt, journal, twoside]{IEEEtran}

\IEEEoverridecommandlockouts                              

\usepackage[hyphens]{url}

\usepackage{graphicx}
\usepackage{subcaption}

\usepackage{amsmath,amsfonts,bm}

\def\eqref#1{equation~\ref{#1}}

\def\1{\bm{1}}

\DeclareMathAlphabet{\mathsfit}{\encodingdefault}{\sfdefault}{m}{sl}
\SetMathAlphabet{\mathsfit}{bold}{\encodingdefault}{\sfdefault}{bx}{n}

\DeclareMathOperator*{\argmax}{arg\,max}

\usepackage{hyperref}       
\usepackage{flushend}

\title{Composing Diverse Policies for Temporally Extended Tasks}

\author{Daniel Angelov, Yordan Hristov, Michael Burke and Subramanian Ramamoorthy%
\thanks{
This paper was recommended for publication by Editor Nancy Amato upon evaluation of the Associate Editor and Reviewers' comments. 
This research is supported by the Engineering and Physical Sciences Research Council (EPSRC), as part of the CDT in Robotics and Autonomous Systems at Heriot-Watt University and The University of Edinburgh. Grant reference EP/L016834/1., and by an Alan Turing Institute sponsored project on Safe AI for Surgical Assistance.
}
\thanks{D. Angelov, Y. Hristov, M. Burke, S. Ramamoorthy are with Institute of Perception, Action and Behaviour (IPAB), School of Informatics, The University of Edinburgh, EH8 9AB, UK;
{\tt\small \{d.angelov, yordan.hristov, michael.burke, s.ramamoorthy\}@ed.ac.uk}}%
\thanks{© 2020 IEEE.  Personal use of this material is permitted.  Permission from IEEE must be obtained for all other uses, in any current or future media, including reprinting/republishing this material for advertising or promotional purposes, creating new collective works, for resale or redistribution to servers or lists, or reuse of any copyrighted component of this work in other works.}
}

\begin{document}

\maketitle
 \thispagestyle{empty}
 \pagestyle{empty}

\begin{abstract}

Robot control policies for temporally extended and sequenced tasks are often characterized by discontinuous switches between different local dynamics. These change-points are often exploited in hierarchical motion planning to build approximate models and to facilitate the design of local, region-specific controllers. 
However, it becomes combinatorially challenging to implement such a pipeline for complex temporally extended tasks, especially when the sub-controllers work on different information streams, time scales and action spaces. In this paper, we introduce a method that can automatically compose diverse policies comprising motion planning trajectories, dynamic motion primitives and neural network controllers. 
We introduce a global goal scoring estimator that uses local, per-motion primitive dynamics models and corresponding activation state-space sets to sequence diverse policies in a locally optimal fashion. 
We use expert demonstrations to convert what is typically viewed as a gradient-based learning process into a planning process without explicitly specifying pre- and post-conditions. 
We first illustrate the proposed framework using an MDP benchmark to showcase robustness to action and model dynamics mismatch, and then with a particularly complex physical gear assembly task, solved on a PR2 robot.
We show that the proposed approach successfully discovers the optimal sequence of controllers and solves both tasks efficiently.

\end{abstract}

\begin{IEEEkeywords}
Motion and Path Planning; Learning and Adaptive Systems; Learning from Demonstration
\end{IEEEkeywords}


 \begin{figure*}[t]
    \centering
    \begin{subfigure}[b]{0.325\linewidth}
        \includegraphics[height=0.93\linewidth]{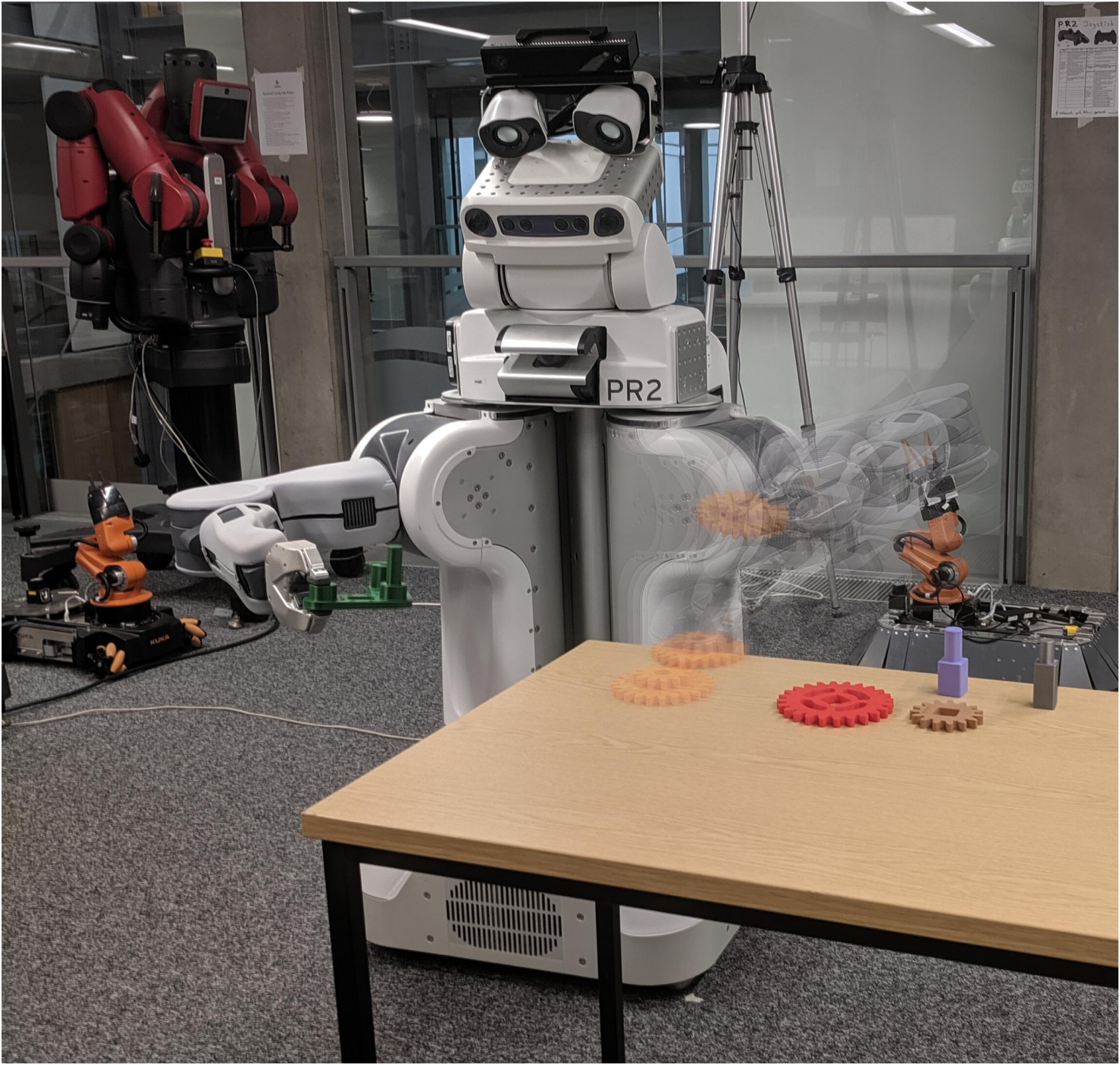}
        \caption{Gear Pick Up}
        \label{fig:pickup}
    \end{subfigure}
    \begin{subfigure}[b]{0.325\linewidth}
        \includegraphics[height=0.93\linewidth]{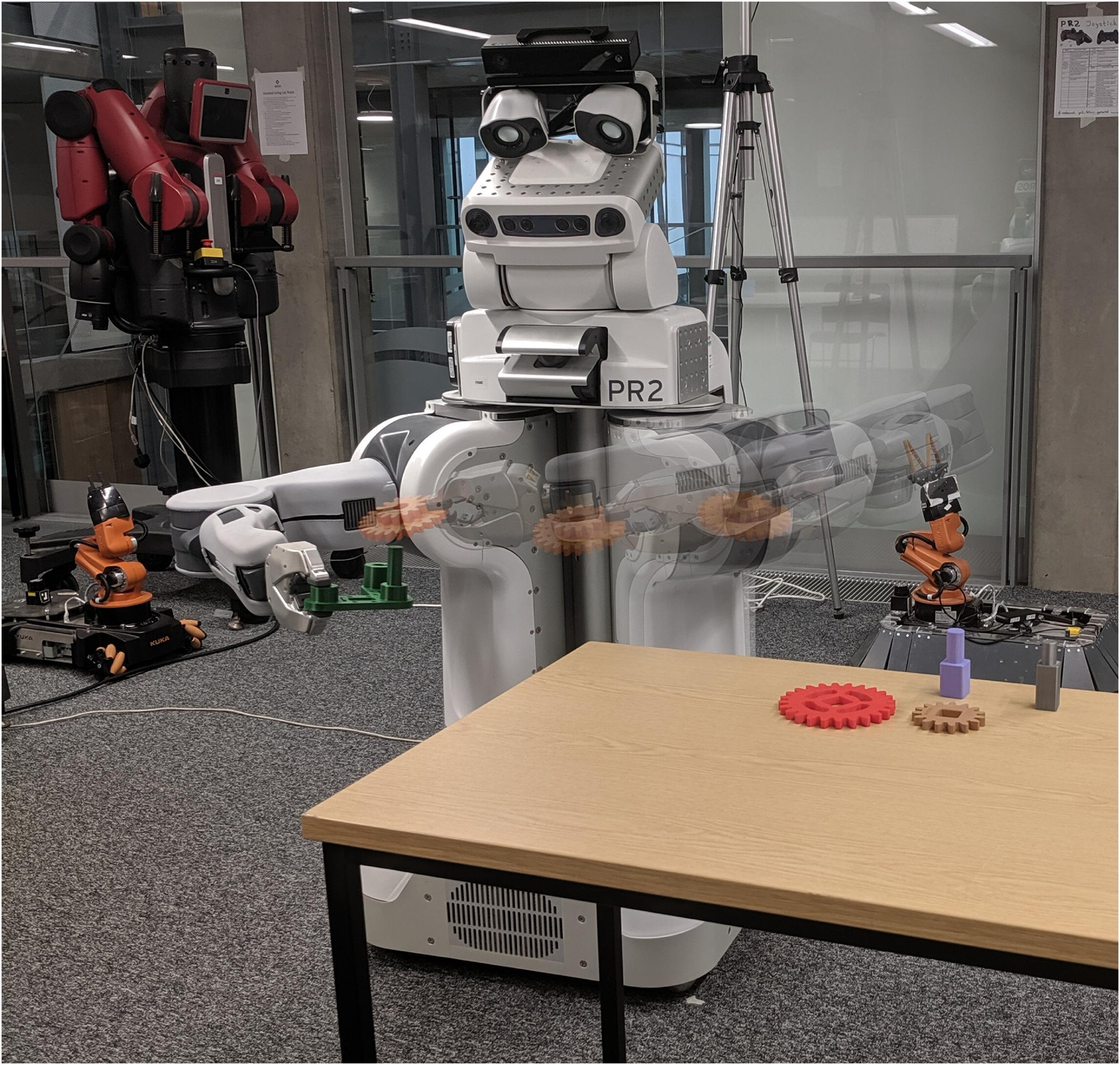}
        \label{fig:move}
        \caption{Move Gear}
    \end{subfigure}
    \begin{subfigure}[b]{0.325\linewidth}
        \includegraphics[height=0.93\linewidth]{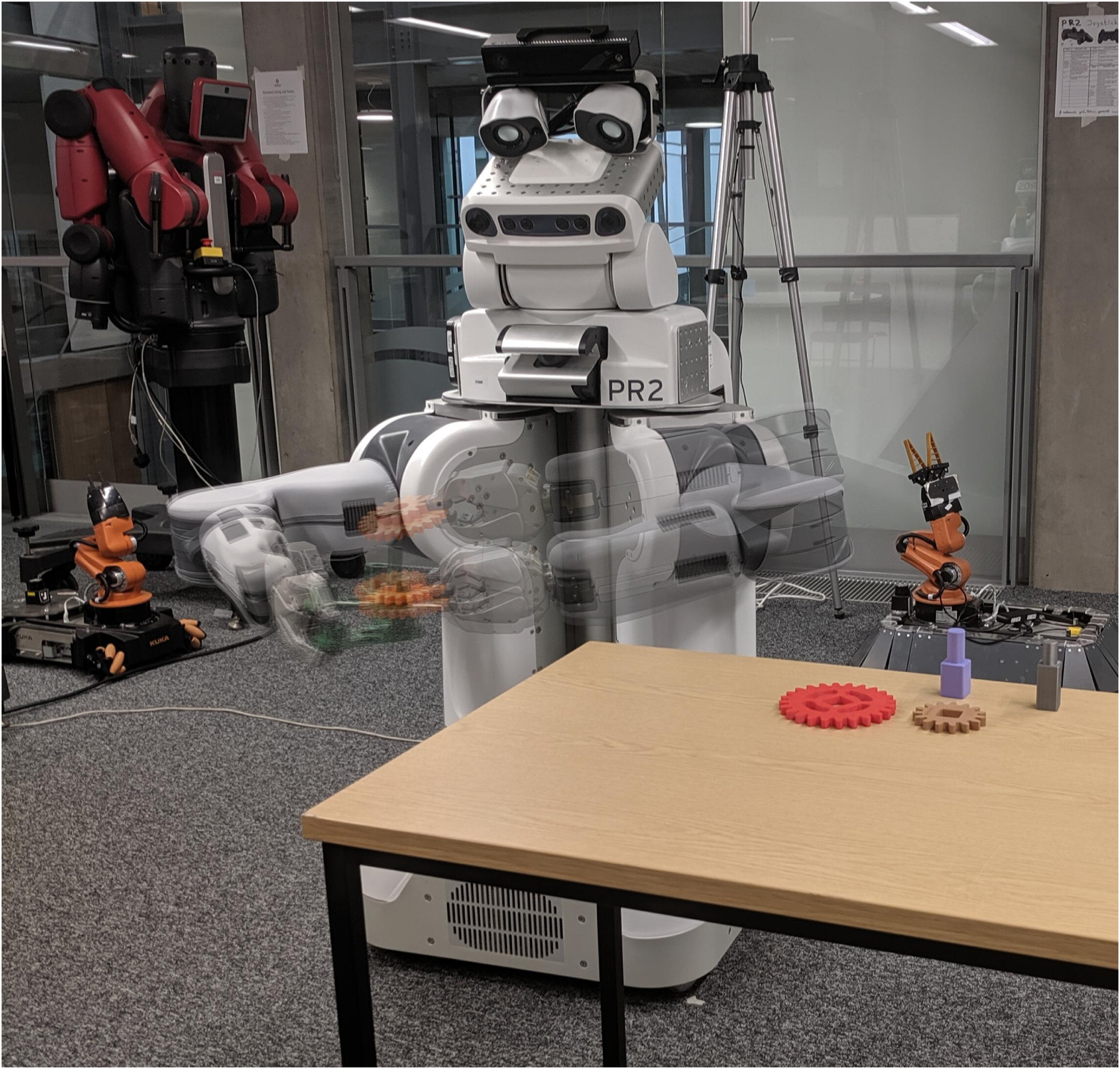}
        \label{fig:insert}
        \caption{Gear Insertion}
    \end{subfigure}
    \caption{Robot setup for the gear assembly task. The robot needs to pick up a gear by leveraging the surface of the table, slide it up to an edge, grasp and move it in a collision free manner to the other hand, before inserting the gear onto the base plate.}
    \label{fig:pr2_task}
\end{figure*}

\section{Introduction}
\IEEEPARstart{F}{or} robots to work in the wild, they need to be able to perform a variety of consecutive tasks that might require vastly different skills.
Each individual skill could be partitioned and optimized outside of this complex system, and is potentially constructed using a number of diverse methods or control strategies, such as motion planning approaches for reaching, contact aware grasping, picking and placing, or through the use of end-to-end neural network based controllers.

In many practical applications, we wish to combine a diversity of such controllers to solve complex tasks. This typically requires that controllers share a common domain representation and a notion of progress to sequence these.
For instance, the problem of assembly, as shown in Figure~\ref{fig:pr2_task}, can be partitioned by first picking up a mechanical part, then using motion planning and trajectory control to move this in close proximity to an assembly, before the subsequent use of a variety of wiggle policies to fit the parts together, as shown by \cite{knepper2013ikeabot}. Alternatively, the policy could be trained in an end-to-end fashion with a neural network, but one may find this difficult for extended tasks with sparse rewards, such as in Figure~\ref{fig:pr2_task}. In the interest of sample efficiency and tractability, such end-to-end learning could be warm-started by using samples from a motion planner, which provides information on how to bring the two pieces together and concentrates effort on learning an alignment policy, as in \cite{thomas2018learning}. Additionally, the completion of these independent sub-tasks can be viewed as a global metric of progress. 

We propose a hybrid hierarchical control strategy that allows for the use of diverse sets of sub-controllers, consisting of commonly used goal-directed motion planning techniques, other strategies such as wiggle, slide and push-against \cite{mason1982manipulator} that are so elegantly used in human manipulation, as well as deep neural network based policies that are represented very differently from their sampling-based motion planning counterparts. 

Thus, we tackle a key challenge associated with existing motion primitive scheduling approaches, which typically assume that a common representation is used by all sub-controllers.
We make use of the fact that controllers tend to have a dynamic model of the active part of their state space - either an analytical or a learned model, and further estimate how close each state is to completing the overall task using a novel goal scoring estimator. 
This allows the hierarchical controller to model the outcome of using any of the available sub-controllers and then determine which of these would bring the world state closest to achieving the desired solution -- in the spirit of model predictive control.

As in the work of \cite{burridge1999sequential} on sequencing funnels and \cite{tedrake10lqrtree} on LQR-Trees, the scheduled controllers for sub-regions of the state space can be optimized in our framework, allowing for compositional task completion, but importantly, also for additional diversity of the controller set. 

Value function approximation techniques used in the reinforcement learning community \cite{kaelbling1996reinforcement} can be considered similar to the proposed progress estimator, but only model the expected reward and require the actions to be in the same state space. We attempt to remedy this oversight, by allowing for a diversity of action and state spaces, and by modelling global progress at a local controller level.

This paper makes the following contributions: 
\begin{itemize}
    \item We use a \textbf{Goal Score estimator} to sequence a set of policies to solve a task. This estimator is trained using expert demonstrations to evaluate the current and future state of the plan and helps to transform the hierarchical learning problem into a \textbf{planning} problem.
    \item We provide a method for composing \textbf{diverse} policies that work with different input information, or decompose the action in either joint or end-effector space and work at different operational frequencies to solve a high level task.
\end{itemize}

We first evaluate the use of the controller dynamics and the goal metric to compose policies in a hybrid controller on an MDP benchmark problem  to evaluate robustness to action and model dynamics noise. Next, we apply this approach to a physical gear assembly task performed by the PR2 robot, making use of both motion planning and visual neural network policies (Figure.~\ref{fig:pr2_task}).

\section{Related Work}

\textbf{Robotics:}
Compositionality is a key paradigm for robot control, which methods of composing controllers of a single type like \cite{burridge1999sequential, tedrake10lqrtree, Burke19Explanation} aim to exploit. These techniques rely on partitioning a state space into smaller overlapping operating regions and tuning sub-controllers (feedback or LQR) for operation in these regions. Unfortunately, these methods often fail to consider the fact that different tasks may require different controller sequences, and the scheduling of control laws in work on compositionality is often underemphasized. Inspired by this capability and the \textit{funnels} framework \cite{mason1995funnel}\footnote{Regions of robustness arising from the dynamics and control applied in a sub-region of the control space.} this work provides a Model Predictive Control (MPC) \cite{garcia1989model} framework for compositional sequencing where controllers can be of different types and operate using different state spaces. 

The ability to act on different state-spaces and action sets is particularly important, as
the sub-policies required to complete a temporally extended task can be highly variable. For example, sub-problems such as grasping and pushing have been addressed and investigated at least since the 1980s, and these could be encapsulated into operation as motion primitives \cite{mason1982manipulator}. Using a diverse set of policies allows for the selection of controllers that best fit the working domain - for example \cite{perez84automatic} highlights that compliance may be needed when movement and sensing reaches the perception noise boundary, \cite{lynch99nonprehensile} advocate using non-prehensile grasps for manipulation of objects and \cite{rodriguez12caging} explore manipulation strategies that allow for caging of objects, such that these can be re-grasped stably in a subsequent stage. Alternatively these motion planning strategies can be formulated using stable nonlinear attractor systems as in DMPs \cite{schaal2006dynamic, ijspeert2013dynamical} or as DeepDMPs \cite{gelada2019deepmdp}.
We aim to create a hybrid control framework that allows the use of these diverse motion planning controllers, alongside neural network policies to solve long-sequence tasks.

\textbf{Learning from Demonstration:}
To expedite the learning process, it is common to provide demonstrated example solution trajectories to a problem.
Methods like Behaviour Cloning (BC) allow for simple visuomotor policies to be learned end-to-end \cite{bojarski2016end}, or to be extended to learn safe policies \cite{huang2018learning}, extract preferences \cite{angelov2018causal} or to learn mappings for the perception and kinematic differences \cite{Argall2009}. Alternatively, they can be used to calculate the relative value of each state through inverse reinforcement learning and to create a hierarchical formulation for control \cite{kolter2008hierarchical}. As explained in \cite{codevilla2019exploring}, there are limitations to BC in terms of number of demonstrations, generalization, and the challenge of modeling complex scenarios. However, we use these full task demonstrations as a means for estimating the distance to a desired goal state, which is arguably a simpler task than learning an entire policy. Additionally, by allowing different controller representations, we do not need to re-represent one control law in alternative approximate forms.

\textbf{Reinforcement Learning:} In the reinforcement learning (RL) literature the concept of options has parallels to our work, as each policy can be viewed as a controller with the initiation set as its domain. Our method lies between learning policies over options as in \cite{barto2003recent}, and computing solutions using learning from demonstration by inverse reinforcement learning \cite{arora2018survey}. 

The options framework \cite{sutton1999between, precup2000eligibility} provides a formal means to work with hierarchically structured sequences of decisions made by a set of RL controllers.
Temporal abstractions have been extensively investigated \cite{drescher1989made, fikes1972learning, iba1989heuristic, korf1983learning, sutton1999between}, and it is clear that hierarchical structure helps to simplify control, allows an observer to disambiguate the different states of the agent, and encapsulates a control policy and termination of the policy within a subset of the state space of the problem. This split in the state space allows us to verify the individual controller within the domain of operation \cite{andonov2015controller, ghosh2018verifying}, deliberate about the cost of an option and increases interpretability \cite{harb2018waiting}. Our work can be viewed as using a planner as a hierarchical policy in the options framework, which is made possible through the incorporation of a goal-scoring progress function learned from demonstration.

In a similar manner, \cite{hafner2018learning} showed how planning can be incorporated into action selection when future states can be evaluated. Our method borrows this view of temporally abstracting trajectories and extends it by applying a dynamics model for each of the options, allowing an agent to assess its states and incorporate foresight \cite{andrychowicz2017hindsight} in its actions.

The work of \cite{mataric1994reward} highlights that including a dense reward indeed increases the overall performance of the agent. Instead of using a predetermined dense function, we learn a Goal Scoring estimator from the demonstrations. As shown in \cite{thomas2018learning} naively tuning and shaping a reward function may result in sub-optimal solutions using base actions. Furthermore, our planner selects an already learned controller and thus avoids converging to sub-optimal behaviours.

As highlighted by Sunderhauf \cite{sunderhauf2018limits}, there are limits of the use of RL in robotics. By leveraging strategies from both RL and control communities, this work aims to increase the scope of problems that can be tackled in robotics.

\section{Method}
\label{sec:def}

Our framework defines a hierarchical controller over the set of pre-existing controllers. Each policy uses its dynamic model to propagate the current state to a future state conditioned on its control law. The Goal Scoring Estimator, learned over expert demonstrations, evaluates those future states and selects a controller that brings the system closest to the desired configuration.

Formally, assume the existence of a learned set of controllers $\mathcal{C}=\{c_1, c_2, ..., c_N\}$ including those learnt from experience in previously solved problems. Using notation similar to the RL options framework \cite{sutton1999between}, each controller $c_\omega$ is independently defined by a control law $\pi_\omega(s) \rightarrow a$, $s \in \mathcal{S}_\omega$, action $a \in \mathcal{A}_\omega$, a working domain $\mathcal{I}_\omega, \mathcal{I}_\omega \subseteq \mathcal{S}_\omega$ where the controller can be started, and a termination criterion $\beta_\omega$. We rely on a forward dynamics model $s_{t+1} \sim \mathcal{D}_\omega(s_t, a_t)$, which is a stochastic mapping, and a Goal Scoring metric $g \sim \mathcal{G}_{K_j}(s_t),~0 \leq g \leq 1$, that estimates the progress of the state $s_t$ with respect to a desired world configuration. We assume $\mathcal{G}_{K_j}$ to change monotonically through the demonstrated trajectories. 
The different controllers can work on different state spaces $\mathcal{S} = \{\mathcal{S}_1, \mathcal{S}_2, .., \mathcal{S}_N\}$ as long as there exists a space $\mathcal{S}^*$, such that $\mathcal{S}_i \subseteq \mathcal{S}^*$. This means there exists a higher or equal order state space, which maps the controller space of operation to regions of $\mathcal{S}^*$.

This work constructs a hybrid hierarchical controller $\pi_\Omega(\omega_t |s_t)$ that can choose the next controller $c_{\omega_t}$ that needs to be executed to bring a learned latent state $s_t$ to some desired $s_{final}$. It uses the forward dynamics model $\mathcal{D}_\omega$ in an $n$-step Model Predictive Control (MPC) look-ahead, using a Goal Scoring metric $\mathcal{G}_K$ that evaluates how close $s_{t+n}$ is to $s_{final}$. 

\begin{figure*}[!h]
    \centering
    \includegraphics[width=1.\linewidth]{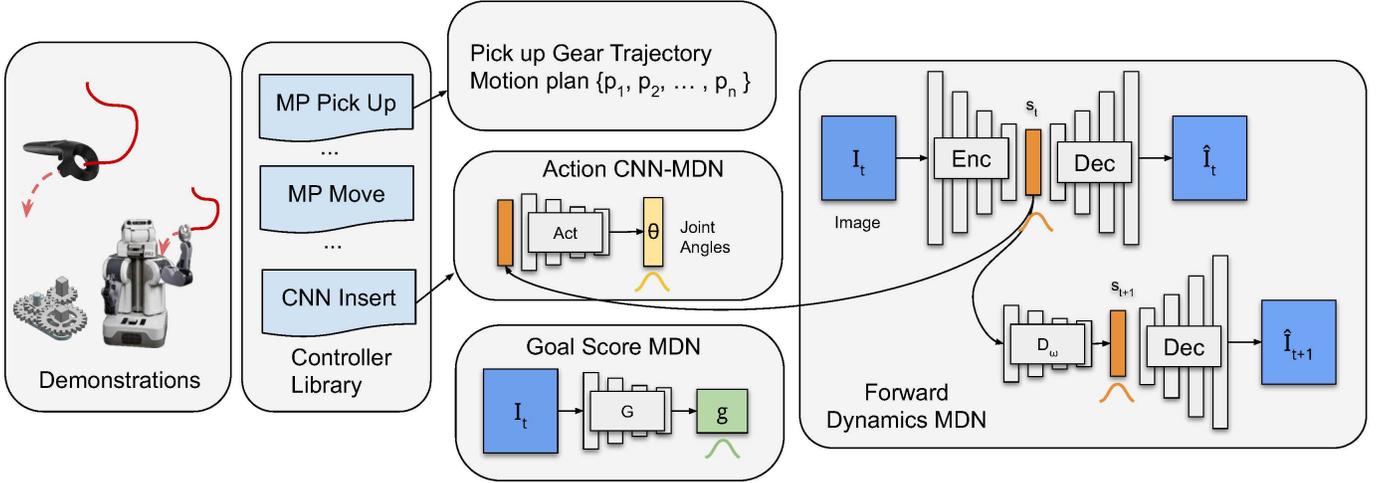}
    \caption{Demonstrations were performed by using an HTC Vive controller that directly teleoperates the end-effector of the PR2 robot at 20 Hz. In the Gear Assembly Task, our controller library includes motion planning (MP) primitives (operating on joint angles) for picking up or moving and a convolutional neural network (CNN) for inserting the gear (operating on images). The MP primitives produce a trajectory for executing a task. The CNN policy takes a latent representation of the image state and generates a distribution over the target joint angles of the robot. The Forward Dynamics models use a VAE representation alongside an $s_{t+1}$ dynamics prediction network that uses the same decoder. The Goal Score Estimator network takes in an image and produces a distribution over how well this image maps to a particular point in the demonstrations.}
    \label{fig:network}
\end{figure*}

As shown in Fig. \ref{fig:network}, in this work, we use a variational autoencoder to learn a latent state $s_t$ from image observations. We assume that each controller in the library has an associated forward dynamics model, trained to predict the next latent state, $s_{t+1}$. This provides us with an implicit mapping between states, and allows us to render an image of an expected scene for each controller that is applied. This scene prediction is then used by the goal score metric to evaluate the effect of choosing each controller and to select the most appropriate controller to be used at a given time step. In effect, this means that controllers act on the appropriate state components, but the underlying state representation used for controller selection is conditioned on image observations. Conditioning on images is feasible, as the robot head camera provides an overhead view of the entire workspace. While it may be possible to learn a shared state representation or mapping between states, this can be challenging (e.g. mapping from joint angles to images is extremely hard), while learning to predict the next latent state is a much easier task. Each of the framework components is described below.

\subsection{Goal Score Evaluation}
The key component of the proposed framework is the ability to evaluate how well a particular state $s$ maps to parts of a demonstrated expert trajectory. This allows us to estimate the temporal distance of that state to the end of the demonstration (see Figure~\ref{fig:network}). 
In a similar manner to \cite{sermanet2018time}, who use adjacency of frames as positive and negative examples, we leverage the temporal sequence of the demonstration as a measure of task completeness.

We capture demonstrations of the global task (in its entirety) to use as a weak supervision for learning a goal scoring network that allows us to map a state to a progress estimation value $g \sim \mathcal{G}(s_t)$ for a given task.
To build the Goal Scoring models, we use a convolutional network head with a Mixture Density Network (MDN) tail to encode the different goal representations based on image observations. The network predicts a distribution over the proximity of the current state to the desired goal state.

The first observation of a demonstration can be viewed as score $0$ -- far away from the goal state, whereas the final observation as score $1.0$ -- a target representation of the world. Even though there may not be a one-to-one mapping between the values within several demonstrations, we rely on the variability in their lengths being encoded within the different modes of the MDN of the Goal Scoring Model.

\subsection{Controller Selection}

At a particular point at state $s_t, s_t \in S^*$ when $c_\omega$ is active, we can compute the goodness of following the current controller given these conditions up to a particular time horizon. The action given by the policy is $a_t=\pi_\omega(\hat{s_t}), \hat{s} \in S_\omega$, and following the dynamics model we can write that:
\begin{align}
s_{t+1}=\mathcal{D}_\omega(s_t, a_t)=\mathcal{D}_\omega(s_t, \pi_\omega(\hat{s_t})). 
\end{align}
As the dynamics model is conditioned on the controller $c_\omega$, we can simplify to $s_{t+1}=\mathcal{D}_\omega(s_t)$.
Chaining this for $n$ steps into the future we obtain 
\begin{align}
s_{t+n}=\mathcal{D}_\omega \circ \mathcal{D}_\omega  \circ \dots \circ \mathcal{D}_\omega(s_t) =\mathcal{D}^n_{\omega}(s_t).     
\end{align}
We can evaluate this future state as \(g_{t+n}= \mathcal{G} \circ  \mathcal{D}^n_\omega(s_t)\).
Thus, the hierarchical controller over controllers can be sequentially optimized,
\begin{align}
    \pi_\Omega(\omega_t|s_t)=\argmax _\omega \left(\mathbb{E}\left[\1_{\mathcal{I}_\omega} (s_t) \cdot \mathcal{G} \circ \mathcal{D}^n_\omega (s_t) \right] \right)
\end{align}
This chooses the controller that is within the operation domain for the current state and delivers the largest goal score estimate after $n$ steps.
After choosing and evaluating the optimal $\pi_\Omega$ with respect to the above criterion, another controller can be selected at the next time step, with repetition until the goal is reached.

\subsection{Controller Dynamics Modelling}
The dynamics of each controller is modelled individually only within its operational domain. This simplifies the complexity the dynamics model has to learn and thus requires less data. Here, we learn a neural dynamics model for each controller that predicts the latent state configuration $s_{t+1}$ from $s_t$, as in \cite{ha2018world}. The architecture shown in Fig. ~\ref{fig:network} is based on a VAE encoding, but includes an additional dynamics network, which predicts the next latent state if a given controller were applied. The same decoder is used to force the two representations not to diverge.

A diverse dynamics network can be used as a prior for each controller \cite{du2019task} and the execution of the controllers themselves can be used to build an individual model using the image state space if it is not provided internally.

\section{Experimental Setup}
\label{sec:exp}

We perform two sets of experiments to investigate the efficacy of the structured hierarchical policy by performing MPC future predictions at each step on a simulated MDP problem and on a much more complex physical gear assembly task on the PR2 robot.

\subsection{Simulated MDP}

\begin{figure}[ht]
\centering
\includegraphics[width=1.0\linewidth]{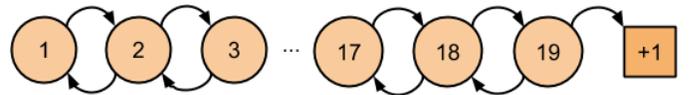}
\caption{The 19-state MDP problem. The action space of the MDP is to move ``left'' or ``right''. The goal of the MDP problem is to reach past state 19 and obtain the +1 reward, which is equivalent to a termination state 20.}
\label{fig:mdp_problem}
\end{figure}

In the first experiment, we use the standard 19-state random walk task as defined in \cite{harutyunyan2018learning} and shown in Figure~\ref{fig:mdp_problem} to illustrate concepts in a simple sequential decision making task.
The goal of the agent is to reach past the $19^{th}$ state and obtain the $+1$ reward. The action space of the agent is to go ``left'' or ``right'', moving the agent to an increasing or decreasing state. There also exist 5 controllers defined as in Section~\ref{sec:def}, with the following policies: (1-3) policies that go ``right'' with a different termination probabilities $\beta = \{0.9, 0.5, 0.2\}$; (4) random action; (5) policy with action to go ``'left'' with $\beta=0.5$. We assume that there exists a noisy dynamics model $\mathcal{D}_\omega$ and the goal evaluation model $\mathcal{G}_{MDP}$, which has the probability of falsely predicting the current state or its value of $0.2$.

Further, we expand the MDP to be of size 100 and evaluate how sensitive the performance of the model is in regards to noise in the Goal Scoring evaluator and in each of the dynamics models.

\subsection{Gear Assembly}
In this task the PR2 robot needs to assemble the first part of the Siemens Challenge\footnote{The challenge is at \href{https://new.siemens.com/us/en/company/fairs-events/robot-learning.html}{https://new.siemens.com/us/en/company/fairs-event}}, which involves grasping a compound gear from a table, and placing it on a peg module held in the other hand of the robot. We record expert demonstrations of the task being performed, and assume access to a set of controllers that (1) picks up the gear from the table; (2) moves the left PR2 arm in proximity to the other arm; (3) inserts the gear on the peg module. 
Policy (1, 2) rely exclusively on scripted path planning techniques and work using discrete time steps, while (3) is learned entirely with a neural network. Controllers (1, 2) share a common state space of the robot's joint angles, whereas (3) works directly on the visual pixel input from the robot's head camera.

The visual neural policy, shown in Figure~\ref{fig:network}, performs imitation learning by using behaviour cloning of the 50 tele-operated demonstrations. This is trained until convergence or 100 epochs using different encoder heads - small convolutional network, ResNet-50, -101. The expert-illustrated trajectories were performed using a HTC Vive controller teleoperating the PR2 robot and the process took less than 1h wall time. The action generation part of the network is an MDN that predicts a distribution of the next time step joint angles $\theta$, which are set as the internal PID targets for the robot 7-DOF arm.  

The dynamics model for each controller is learned independently and is represented with a Forward Dynamics MDN, learned from forward rollouts of the policy network. The Goal Score estimator is learned on an additional 5 rollouts of the full gear assembly task and operates on the latent space of the particular policy. Throughout all of the experiments we use the Adam optimizer with a weight decay rate of $1e^-6$, batch size of $120$, train for $200$ epoch and the MDN uses $24$ Gaussian mixtures. We show the performance of this model with several video streams from different cameras on the robot (head, left and right forearm cameras).

Additionally, we compare the performance of the scripted Motion Planning method (using RRT Connect \cite{kuffner2000rrt}), Dynamic Motion Primitives (learned from the MPs) and the Visual Neural Policy on each subtask, as well as using the full sequence under the different controllers as a baseline.

\begin{figure}[h]
    \centering
    \includegraphics[width=1.\linewidth]{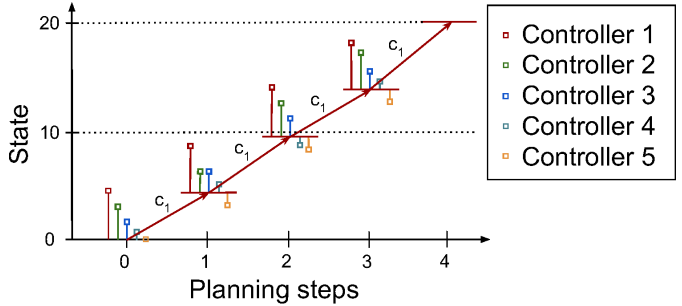} %
    \caption{MDP solution. At timestep 0, a rollout of the 5 controllers is performed with the dynamics model. The expected resulting state is marked using vertical bars. The best performing controller is used within the environment to obtain the next state - the red line at state 5 and planning step 1. This process is iterated until a desired state is reached.}
    \label{fig:mdp_solution}
\end{figure}

\section{Experimental Results}
\label{sec:result}

We demonstrate the viability of composing diverse policies by using the controller dynamics as a method for choosing a satisfactory policy. The dynamics can be learned independently of the task, and can be used to solve a downstream task.

\textbf{Simulated MDP}~~ This problem illustrates the feasibility of using our architecture as a planning method. Figure~\ref{fig:mdp_solution} shows that the agent reaches the optimal state in just 4 planning steps, where each planning step is a rollout of a controller. The predicted state under the specified time horizon is illustrated at each step for the different controller options. This naturally suggests the use of the policy $\pi_1$ that outperforms the alternatives ($\pi_1$ reaches state 6, $\pi_2$ - state 4, $\pi_2$ - state 3, $\pi_3$ - state 1, $\pi_4$ - state 1, $\pi_5$ - state 0). Even though the predicted state differs from the true rollout of the policy, it allows the hierarchical controller to use the controller that would progress the state the furthest. The execution of some controllers (i.e. $c_5$ in planning steps 1, 2, 3) reverts the state of the world to a less desirable one. By using the forward dynamics, we can avoid sampling these undesirable controllers. 

\begin{figure}[h]
    \centering
    \includegraphics[width=.7\linewidth, trim = {1.5cm 0.3cm 0.25cm .4cm}, clip]{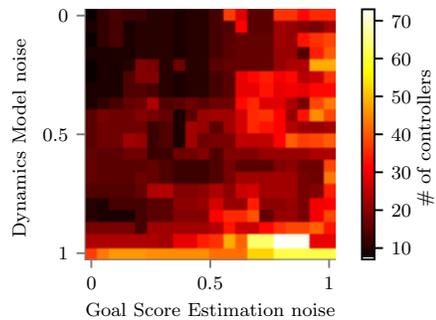}
    \caption{Sensitivity to noise in the dynamics model and the Goal Score Estimator for a world of size 100. The heatmap illustrates the number of controllers that were used in order to reach the target with a lower number - top left - being optimal. The number of controllers varies between the optimal 8 and 72.}
    \label{fig:mdp_noise}
\end{figure}

\begin{figure*}[t]
\centering
    \begin{subfigure}[b]{0.245\linewidth}
        \includegraphics[height=0.98\linewidth]{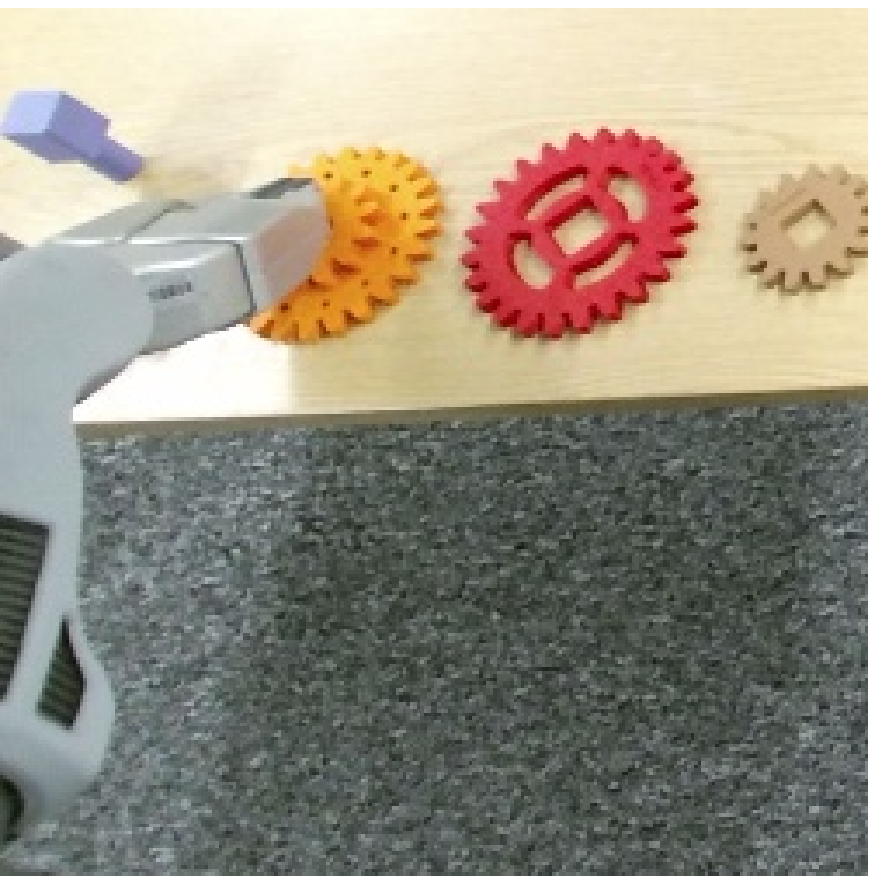}
        \caption{Contact gear}
    \end{subfigure}
    \begin{subfigure}[b]{0.245\linewidth}
        \includegraphics[height=0.98\linewidth]{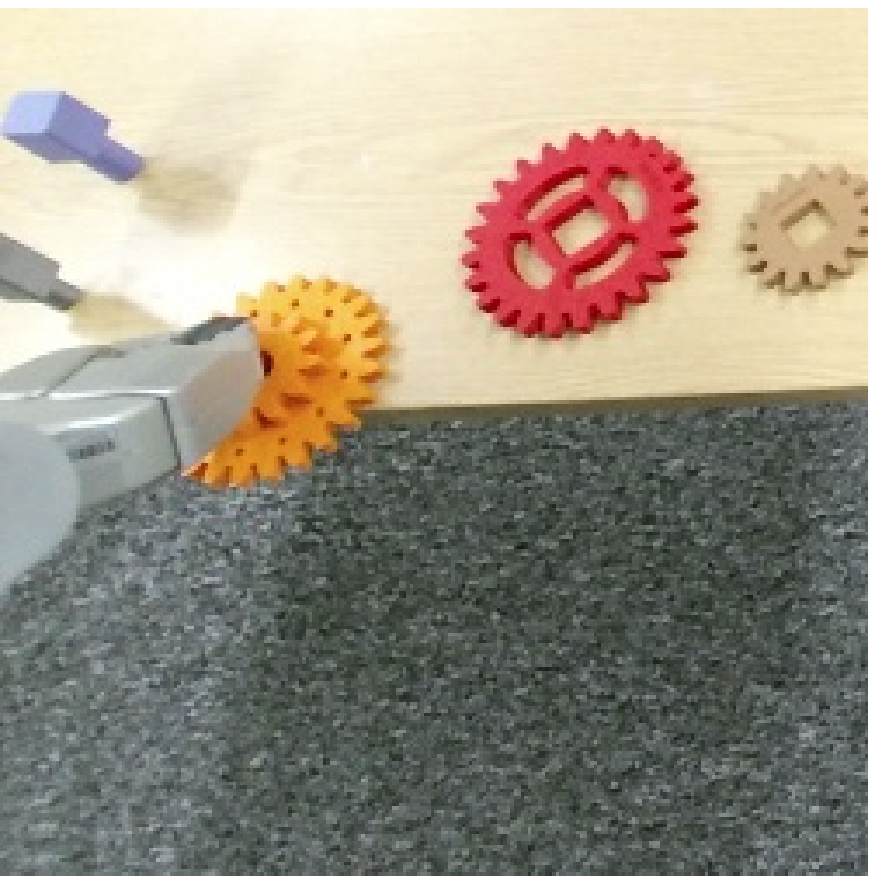}
        \caption{Slide along table}
    \end{subfigure}
    \begin{subfigure}[b]{0.245\linewidth}
        \includegraphics[height=0.98\linewidth]{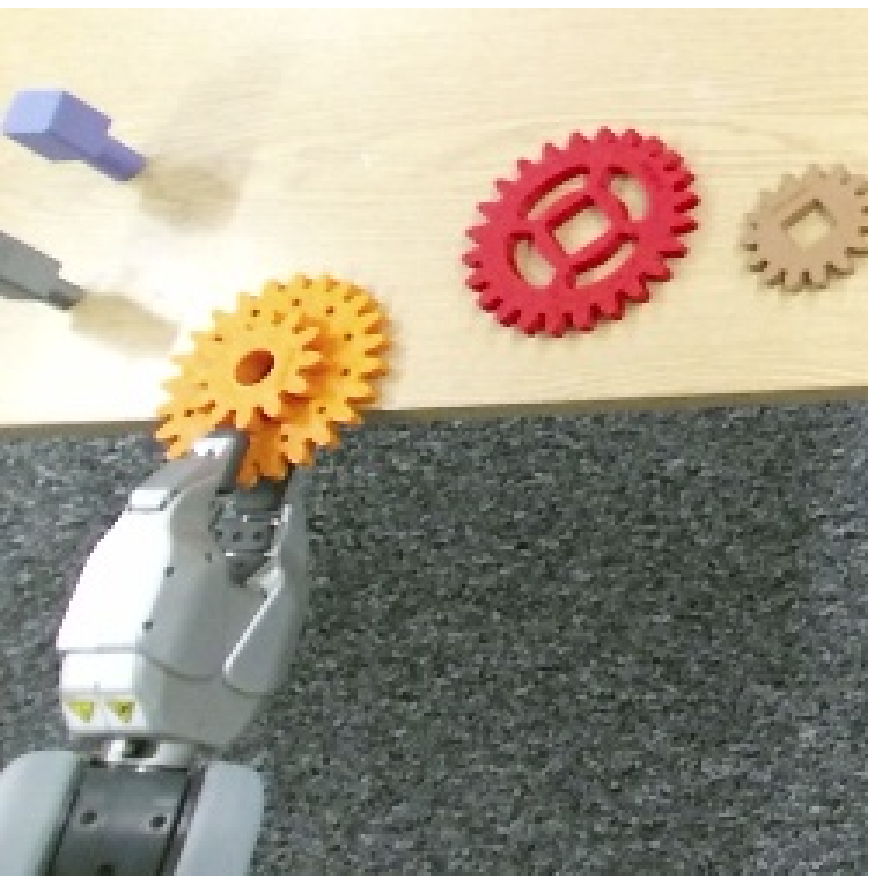}
        \caption{Re-grasp stably}
    \end{subfigure}
    \begin{subfigure}[b]{0.245\linewidth}
        \includegraphics[height=0.98\linewidth]{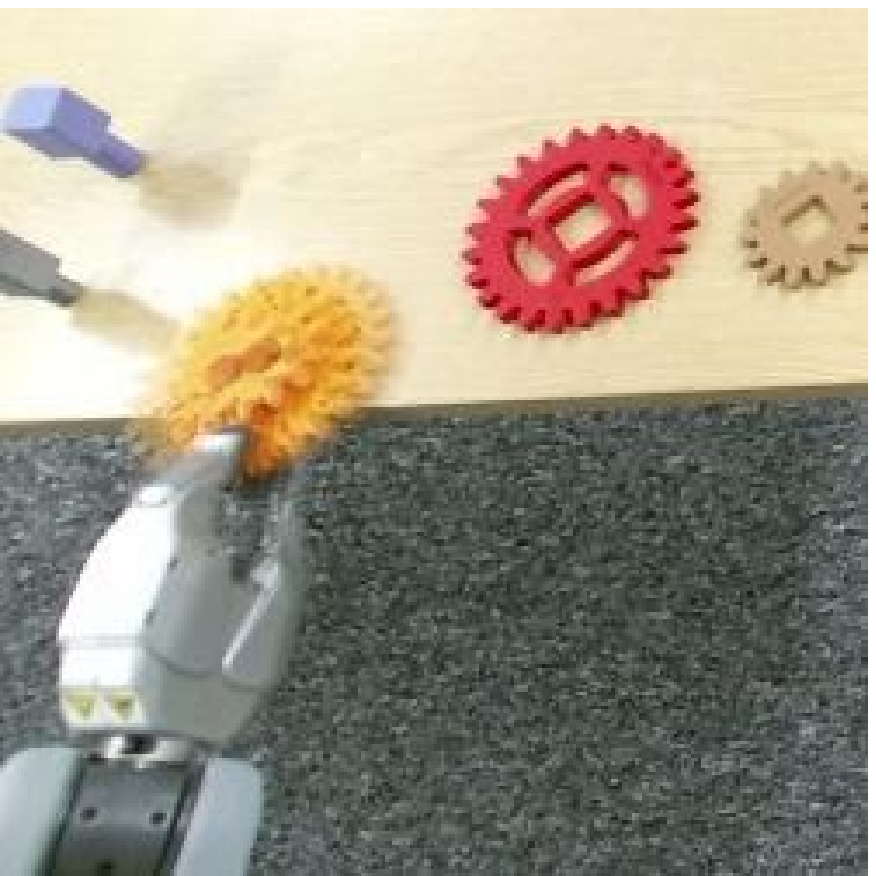}
        \label{fig:grasp_var}
        \caption{Grasp variability}
    \end{subfigure}
\caption{Images \textbf{(a-c)} illustrate key frames of the pick up policy that involves making physical contact with the gear, sliding it along the table surface to an edge and grasping it firmly in the new position. \textbf{(d)} A visual overlay of 3 random pickup attempts. The difference in grasp position relative to the gear is comparable to the inner diameter and is a byproduct of the stochasticity in the sliding and grasping action. This does not hinder the performance of the CNN in the full task.}
\label{fig:gear_pickup}
\end{figure*}

In order to investigate the robustness and convergence properties of our method, we introduce noise within the system, while expanding the MDP to be of size 100 and maintaining the same 5 controllers as above. We can see in Figure.~\ref{fig:mdp_noise} how the number of controllers required to reach the target location varies at different noise levels. When we observe low amounts of noise, the performance remains stable and requires activating any of these controllers a total of less than 20 times (top-left part of the heatmap). The expected optimal number of controller activations based on policy 1 is 12 (black region of the heatmap). As the noise in both the dynamics model and the Goal Score Estimation increases, we observe a degradation and the selection of more sub-optimal controllers. The model is more sensitive to noise in the Goal Score Estimator than when the dynamics of the controllers make errors in their predictions. Despite this, the method converges to the optimal state.

It is interesting to note that the method uses close to or optimal number of controller activations in cases where multiple policies would drive the world in a progressive state, highlighting that the goal score metric is capable of choosing longer horizon controllers due to the MPC look ahead.

\begin{figure}[h]
\centering
\includegraphics[width=1.\linewidth]{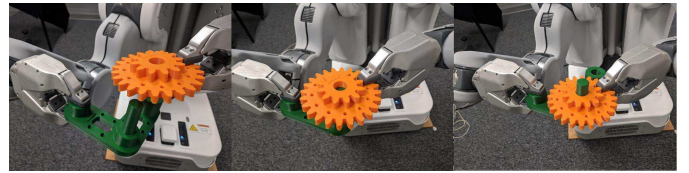}
\caption{The execution of a neural network policy for inserting the gear on the peg. }
\label{fig:gear_peg}
\end{figure}

\textbf{Gear Assembly}~~ We build the library of controllers for the task - picking up a gear (Figure ~\ref{fig:gear_pickup}), moving it close to the base of the assembly and inserting the gear on the base plate (Figure.~\ref{fig:gear_peg}). A motion planning control method was used to perform different tasks. Those demonstrations were used to build the DMP model, using the ROS-DMP module, which is based on \cite{ijspeert2013dynamical}. The Convolutional Neural Network policy was trained using 50 tele-operated demonstrations covering a wide variety of initialization cases for each specific task. We did not observe any task performance changes between the small Convolutional or the ResNet-50,-101 head and therefore relied on the simple architecture. Other tasks may benefit from deeper or more complex models (such as \cite{vecerik2019practical, zeng2019tossingbot, yu2018one}), but integration within the method would remain the same.

\begin{table}[h]
    \caption{Table of successful trials for different policies. MP - Motion Planning, DMP - Dynamic Motion Primitive, CNN - Convolutional Neural Network. The CNN policy has a maximum of 50 steps to reach the goal. The symbol * indicates policies terminated early due to safety concerns.}
    \centering
    \begin{tabular}{lcccc}
    \hline
    Control Method & Pick Up & Gear Move & Gear Insert & Full Task \\ \hline
    MP     & 10/10        & 10/10     & 1/10        & 1/10      \\
    DMP    & 10/10        & 10/10     & 1/10        & 1/10      \\
    CNN    & *            & 10/10     & 10/10       & *      \\ \hline
    \textbf{MP~\&~CNN} (Our) & 10/10        & 10/10     & 10/10       & \textbf{10/10}     \\ \hline
    \end{tabular}
    \label{tab:option_success}
\end{table}

Table.~\ref{tab:option_success} shows the performance of the different controllers on different tasks. The MP and DMP models exhibit stable performance in contact based tasks, but fail where the initial conditions differ -- in Figure~\ref{fig:gear_pickup} we can see the variability that the pickup controller exhibits in terms of the location of the grasp on the gear, which leads to failures in attempting to insert this onto the base assembly. The issue comes from the tolerances of the fit as using an MP and a sequence of trajectory points does not compensate for any inaccuracies incurred during the previous stages of the process or manual positioning. Precise insertion is known to fail outside of a very small convergence basin when using MP controllers - we obtain similar (bad) performance similar to \cite{thomas2018learning, luo2019reinforcement}.

As a baseline, we compare against optimally sequencing the MP and DMP control strategies, which can be seen under the ``Full Task'' performance. Due to the low performance on a part of the task, the overall success rate is limited.

In contrast, the natural variability of the grasp is part of the training set of the CNN model and successfully inserts the gear even with a high variance of initial locations (Figure.~\ref{fig:gear_peg}). As the visual CNN policy is not dependant on the absolute position of either the grasped location or the position of the base assembly, it performs corrective/feedback actions for the policy to succeed. However, the CNN performance on the pickup task could not be evaluated, as the prescribed controller actions were jerky and violated safety constraints (pre-defined velocity and position limits). 

 \begin{figure*}[h]
    \centering
    \begin{subfigure}[b]{.325\linewidth}
        \includegraphics[width=\linewidth]{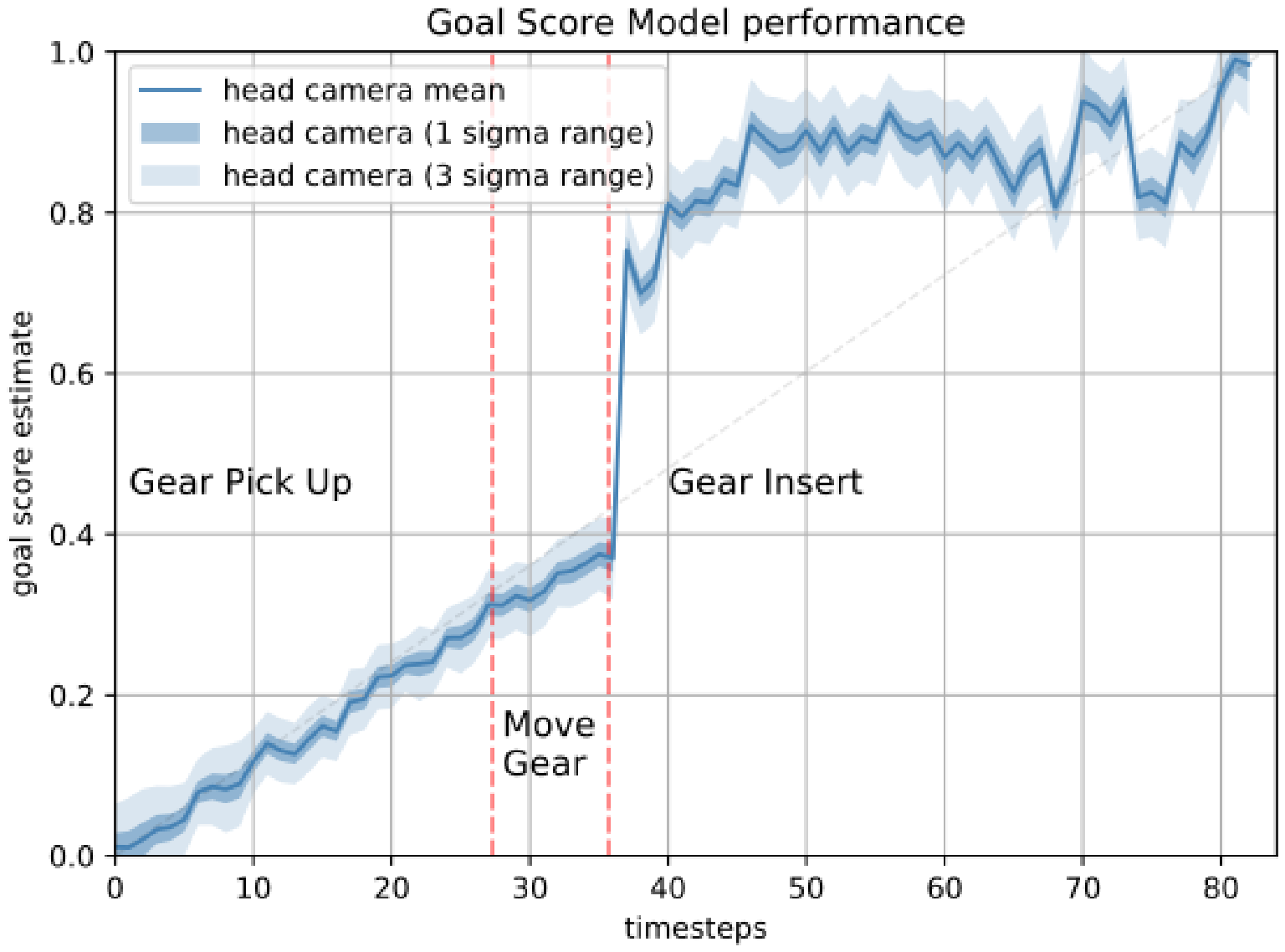}
        \label{fig:head_gsm}
        \caption{Head Camera}
    \end{subfigure}
    \hfill
    \begin{subfigure}[b]{.325\linewidth}
        \includegraphics[width=\linewidth]{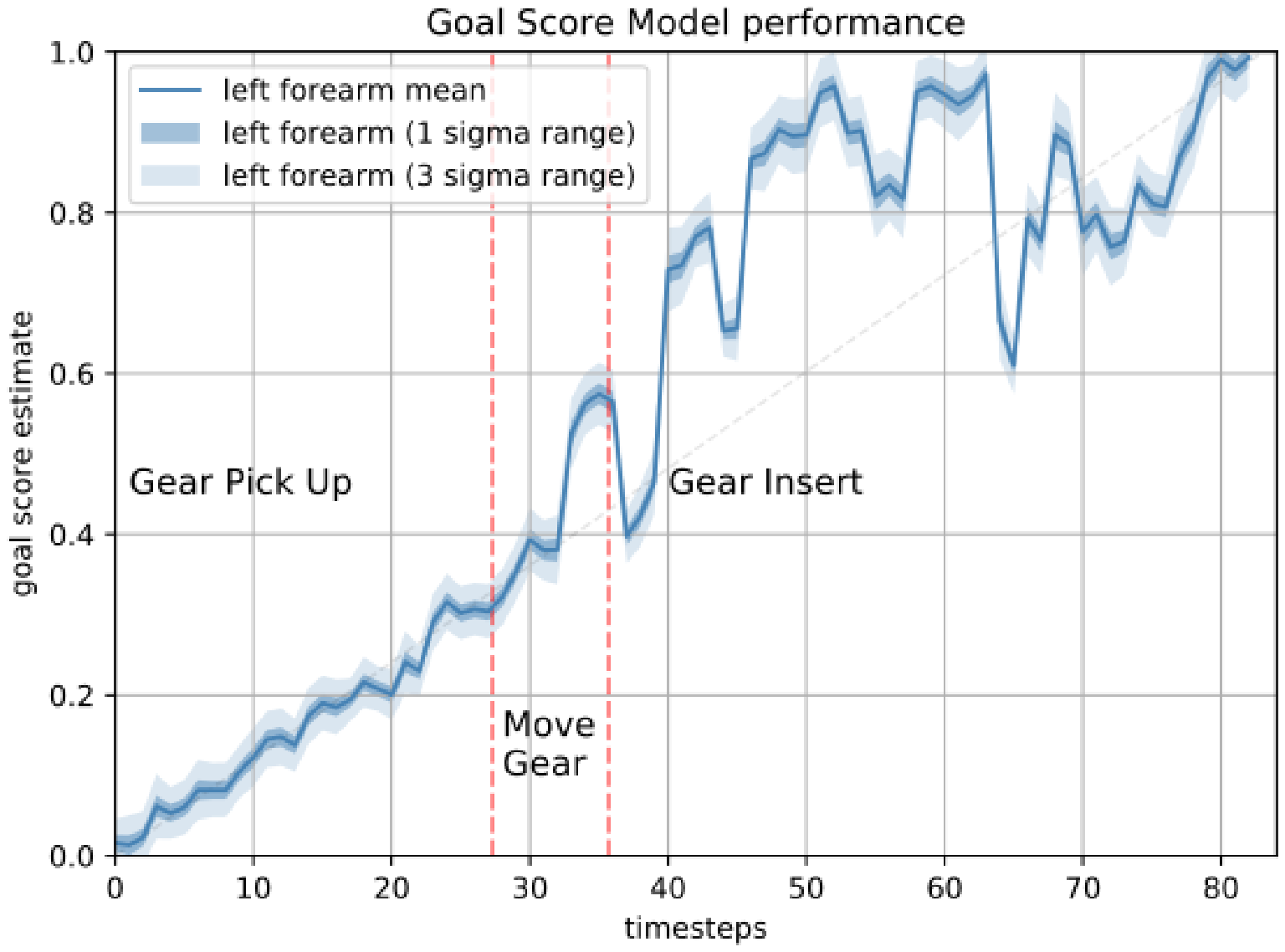}
        \label{fig:left_gsm}
        \caption{Left Forearm}
    \end{subfigure}
    \hfill
    \begin{subfigure}[b]{.325\linewidth}
        \includegraphics[width=\linewidth]{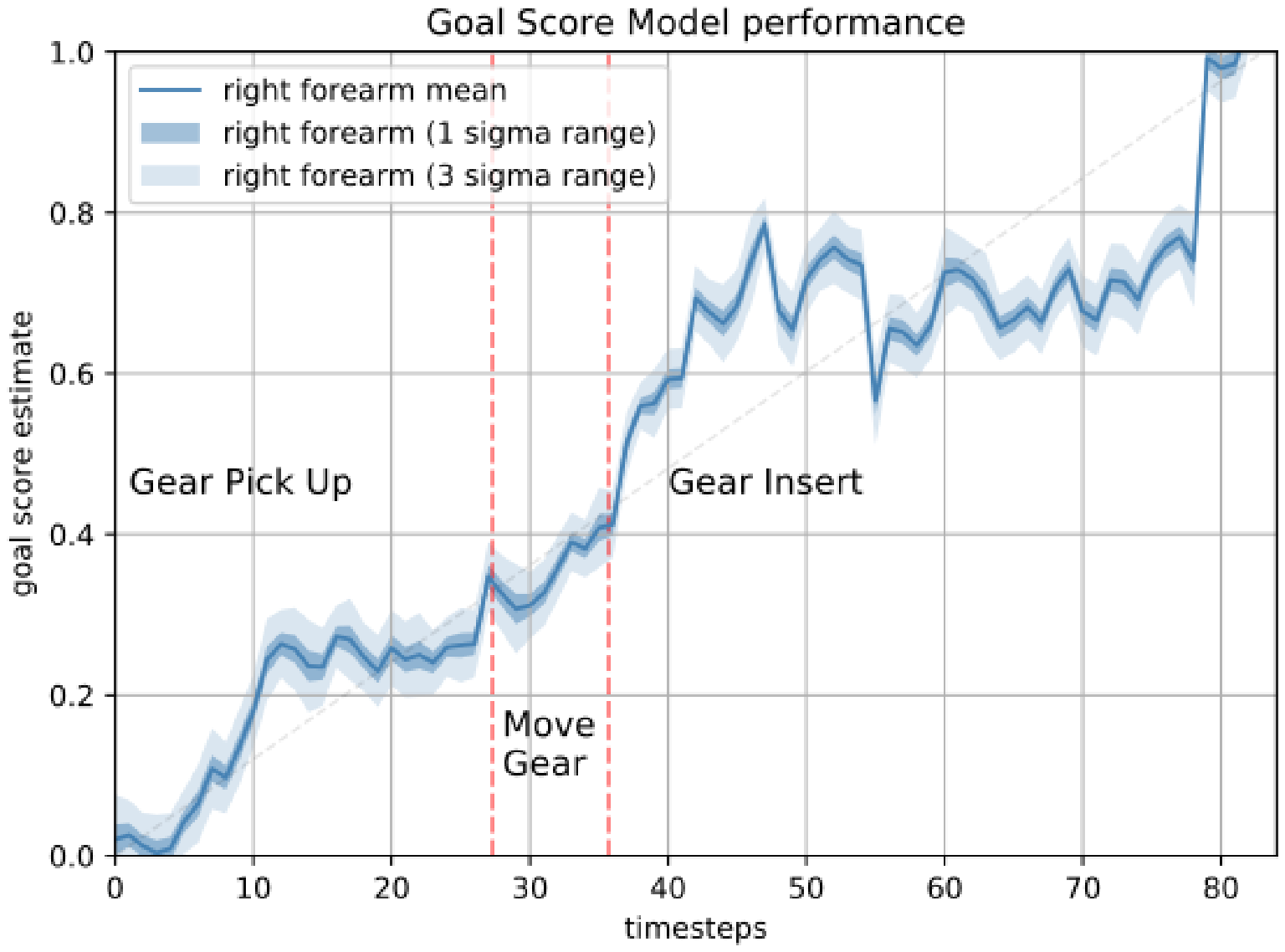}
        \label{fig:right_gsm}
        \caption{Right Forearm}
    \end{subfigure}
    \caption{The goal score metric calculated during the execution of a random trial. During the first two motion planning controllers, the model is monotonically increasing the goal metric. The stochasticity of the neural network policy leads to oscillating scores. Using different input streams, the prediction accuracy could be altered -- the scene head camera does not see the fine details of the movement which the forearm cameras do, leading to a closer to goal score. The peaks in the forearm cameras are associated with states where the peg is extremely close to the gear hole, highlighting that proximity. Example snapshots from different views can be seen in Figure~\ref{fig:views}.}
    \label{fig:goal_score_model_perf}
\end{figure*}

\begin{figure*}[h]
   \centering
    \begin{subfigure}[n]{0.39\linewidth}
        \includegraphics[width=\linewidth]{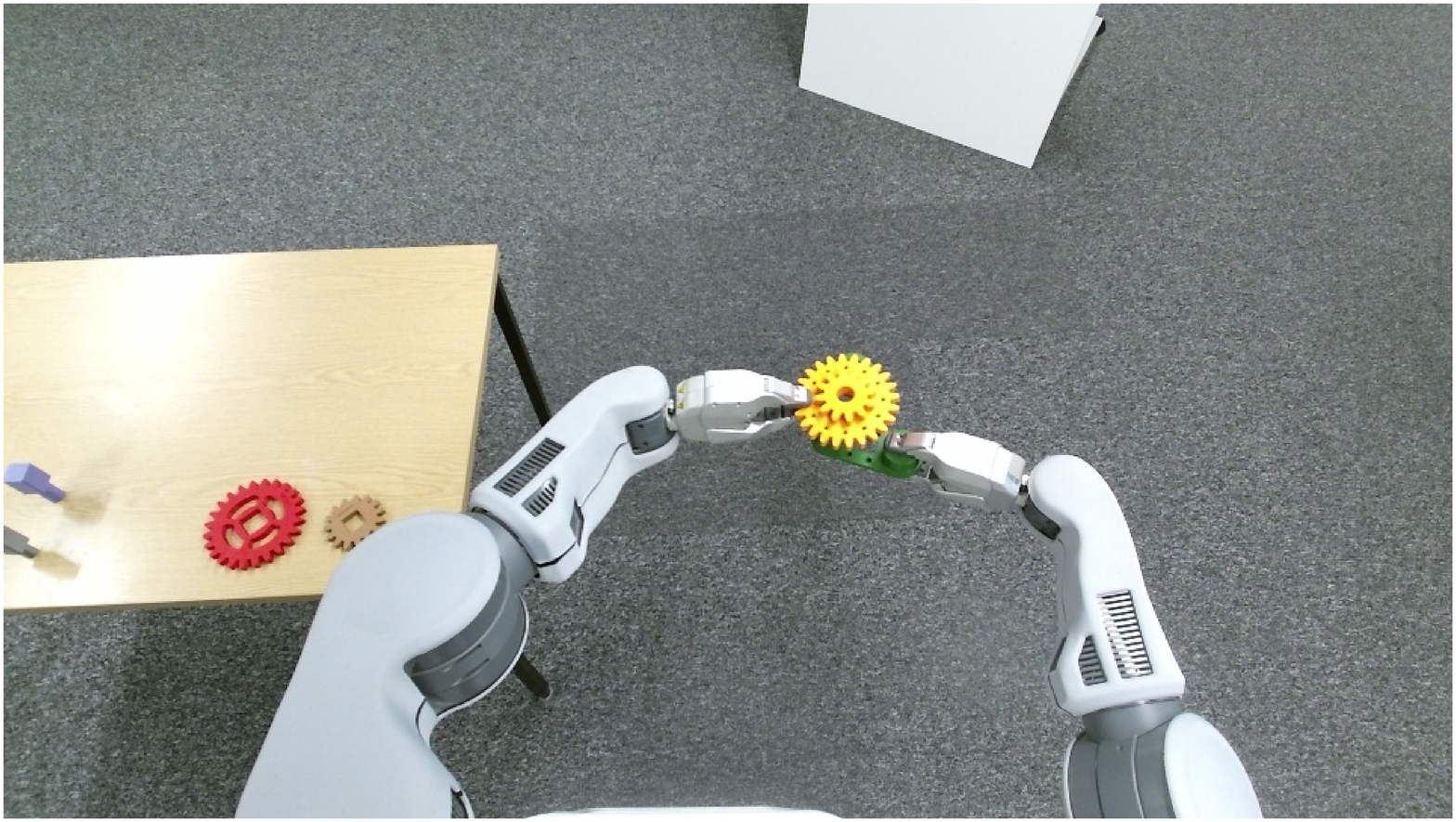}
        \label{fig:head_image}
        \caption{Head Camera}
    \end{subfigure}
    \begin{subfigure}[n]{0.29\linewidth}
        \includegraphics[width=\linewidth]{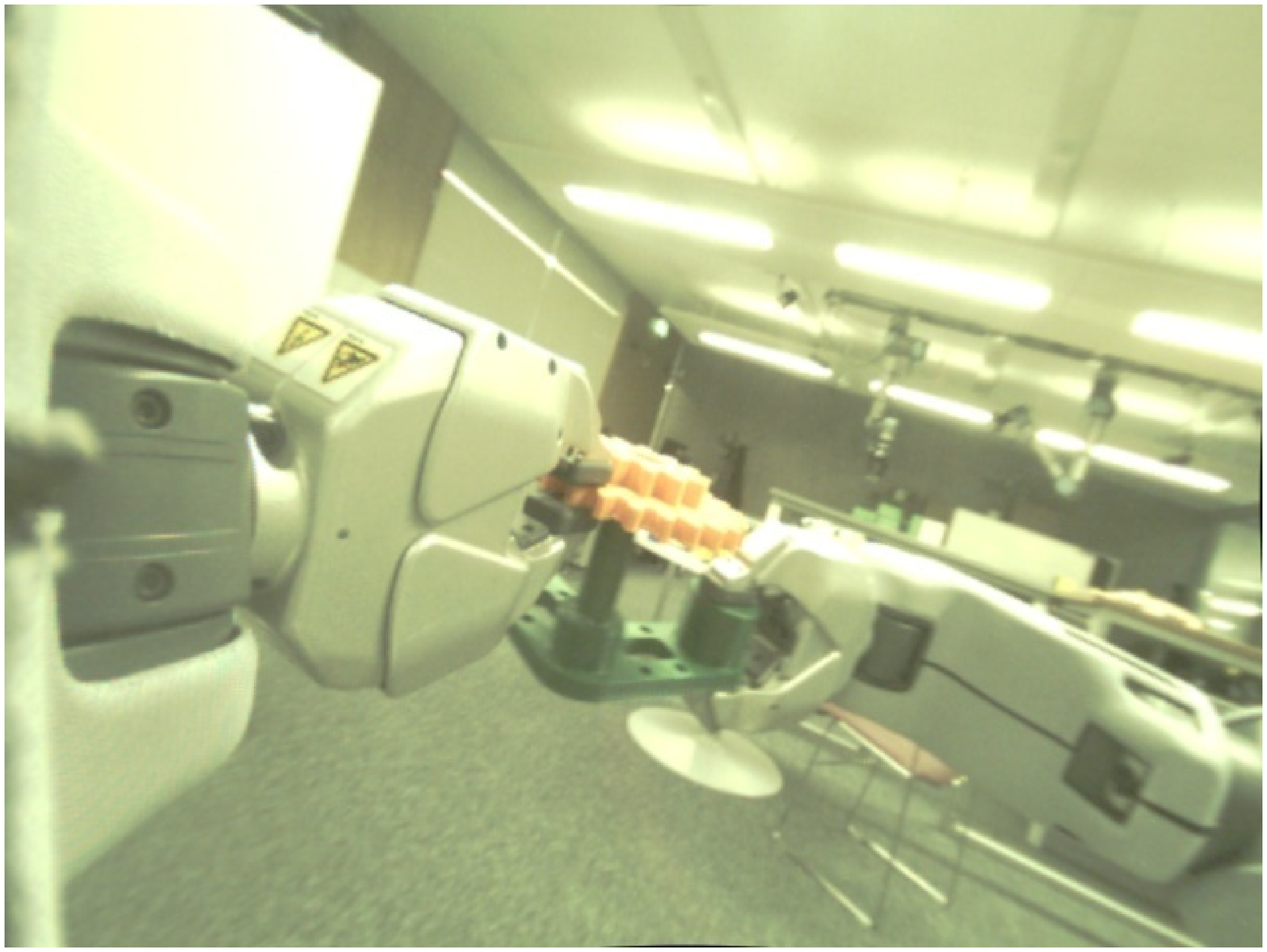}
        \label{fig:left_image}
        \caption{Left Forearm}
    \end{subfigure}
    \begin{subfigure}[n]{0.29\linewidth}
        \includegraphics[width=\linewidth]{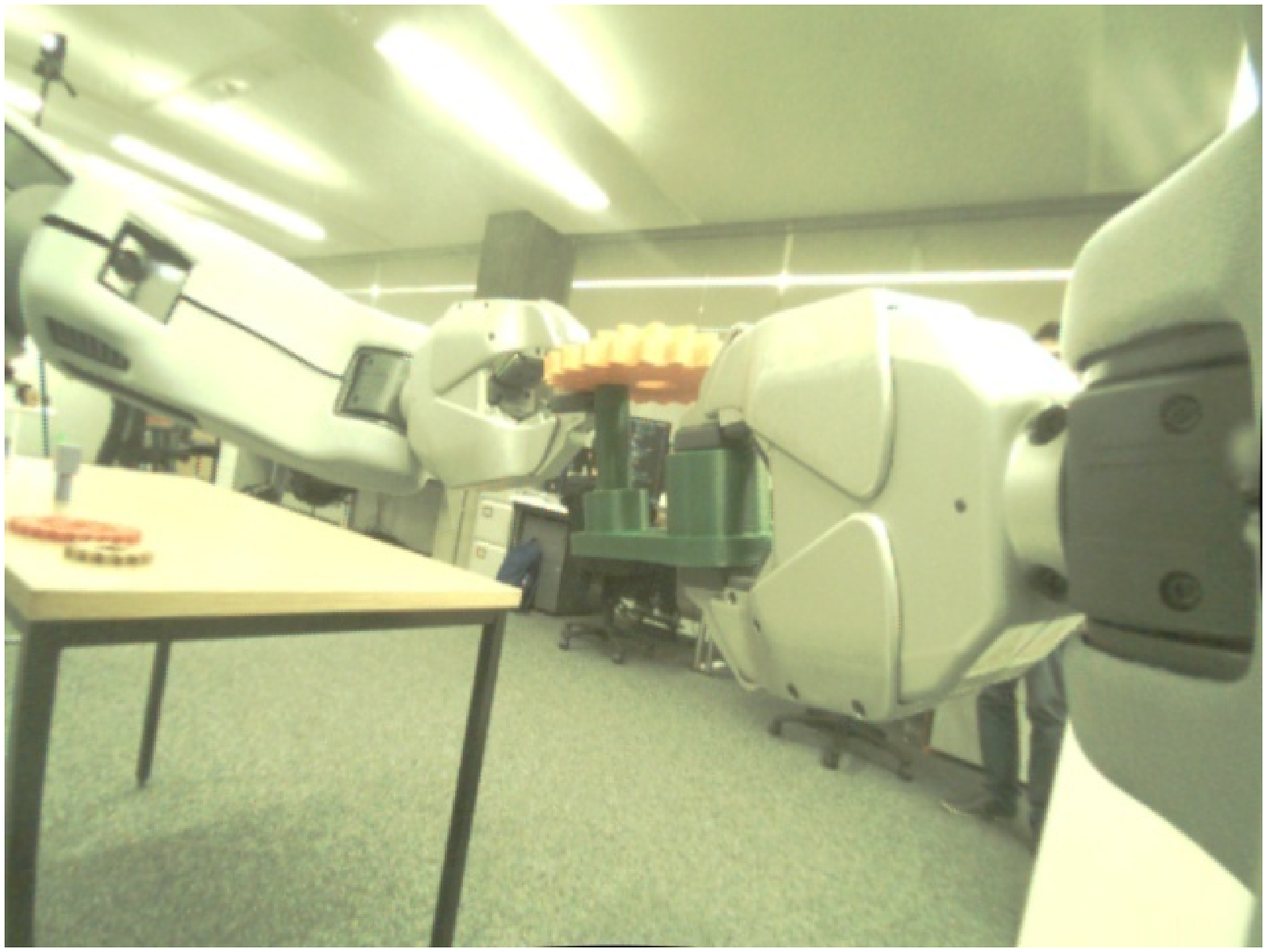}
        \label{fig:right_image}
        \caption{Right Forearm}
    \end{subfigure}
   \caption{Snapshots of the input from different cameras on the PR2 robot demonstrate a moment, where the head camera cannot differentiate how well the task is performed, the left camera is optimistic from its perspective, while the right accurately evaluates the performance as sub-optimal, leading to the goal scoring network predicting a decreased value.}
   \label{fig:views}
\end{figure*}

This illustrates that the combination of \textbf{MP} for picking up the gear and moving it closer to the assembly and \textbf{CNN} to insert the gear, selected using our method allows for the full task to be successfully solved optimally 10 out of the 10 attempts. This shows the advantage of using a diverse set of controllers, allowing each one to be tuned to the domain of operation.

The Goal Score Model is trained on only 5 full task demonstrations. We empirically choose $n=10$ for the $n$ step MPC look ahead as our planning horizon.
In the interests of reproducibility, more information about the sub-controllers and training routines is available on the website \footnote{\url{https://sites.google.com/view/composingdiverse}}.
Figure~\ref{fig:goal_score_model_perf} illustrates the Goal Score estimation for a previously unseen demonstration from camera streams with different viewpoints. The score for the different controllers can clearly be used to sequence the policies. This is shown by the fact that the score follows a monotonically increasing value with regards to the average score for the individual controller domain.

\section{Conclusion}
\label{sec:conclusion}

We introduce a method for composing diverse policies with varied representations, including Motion Planning, Dynamic Motion Primitives and Convolutional Neural Networks. This allows for the solution of combinatorially complex and temporally extended tasks requiring multiple steps, without needing to predefine controller sequences or design high level state machines. We sequence tasks by using a Goal Scoring Model trained by expert demonstrations providing a weak supervisory signal. The goal scoring model provides a controller invariant prediction of progress towards a goal, which can be used with shared latent space across sub-controllers. This work has also introduced different methods that allow for a model-based or a model-free way to create a dynamics model, which can be used to analytically plan the next best option within a model predictive control framework.



\bibliographystyle{unsrt}

\bibliography{main}  


\end{document}